\documentclass{article}
\usepackage{spconf,amsmath,epsfig}

\usepackage{algorithm}
\usepackage{algorithmic}
\usepackage{amsmath}
\usepackage{amsfonts}
\usepackage{arydshln} 
\usepackage[numbers,sort&compress]{natbib} 
\usepackage{enumitem}
\usepackage{caption}  
\usepackage[colorlinks,linkcolor=blue]{hyperref}

\let\OLDthebibliography\thebibliography
\renewcommand\thebibliography[1]{
  \OLDthebibliography{#1}
  \setlength{\parskip}{0pt}
  \setlength{\itemsep}{0pt plus 0.3ex}
}

\begin{document}\sloppy

\def\x{{\mathbf x}}
\def\L{{\cal L}}

\title{MoEController: Instruction-based Arbitrary Image Manipulation with Mixture-of-Expert Controllers}
%
\name{Sijia Li, Chen Chen\sthanks{Corresponding authors.}, Haonan Lu\footnotemark[1]}
\address{OPPO Research Institute, Shenzhen, China}

%

\twocolumn[{
\renewcommand\twocolumn[1][]{#1}
\maketitle
\begin{center}
    \captionsetup{type=figure}
    \includegraphics[width=\textwidth]{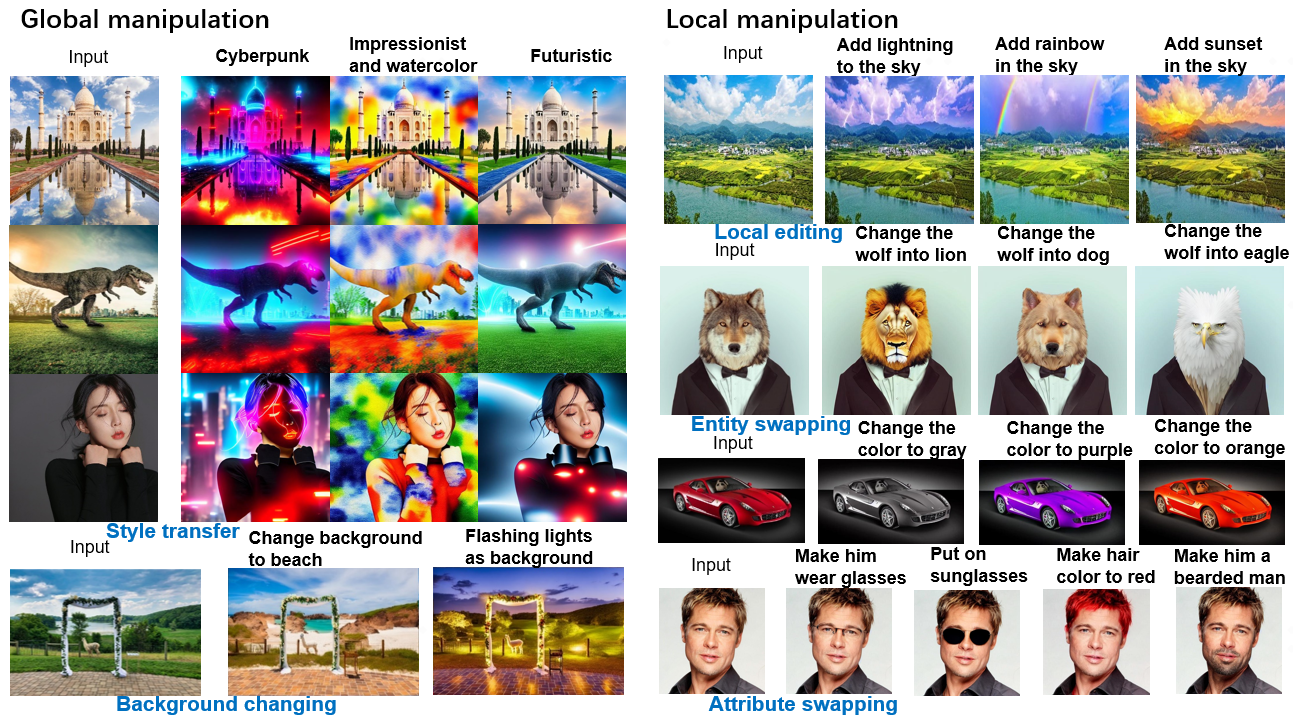}
    \captionof{figure}{Arbitrary instruction-guided image global and local manipulation visualizations.}
\label{fig:1}
\end{center}
}]

\footnotetext[1]{*Corresponding authors(chenchen4@oppo.com; luhaonan@oppo.com)}

%
\begin{abstract}
Diffusion-model-based text-guided image generation has recently made astounding progress, producing fascinating results in open-domain image manipulation tasks. Few models, however, currently have complete zero-shot capabilities for both global and local image editing due to the complexity and diversity of image manipulation tasks. In this work, we propose a method with a mixture-of-expert (MOE) controllers to align the text-guided capacity of diffusion models with different kinds of human instructions, enabling our model to handle various open-domain image manipulation tasks with natural language instructions.
First, we use large language models (ChatGPT) and conditional image synthesis models (ControlNet) to generate a global image transfer dataset in addition to the instruction-based local image editing dataset. Then, using an MOE technique and task-specific adaptation training on a large-scale dataset, our conditional diffusion model can edit images globally and locally.
Extensive experiments demonstrate that our approach performs surprisingly well on various image manipulation tasks when dealing with open-domain images and arbitrary human instructions. 
\end{abstract}
\section{Introduction}
\label{sec:intro}
The ability to manipulate images with text has been demonstrated in earlier works combining GAN and CLIP \cite{patashnik2021styleclip, crowson2022vqgan, couairon2022flexit, kwon2022clipstyler, bar2022text2live}, but the quality and diversity of the generated images have some limitations. Recently, significant progress has been made in these diffusion-based works \cite{meng2021sdedit, hertz2022prompt, tumanyan2023plug, kawar2023imagic, cao2023masactrl, chen2023artfusion, yang2023paint} on image manipulation tasks.
Open-domain image manipulation using only arbitrary text instructions, however, is still a challenging and comprehensive task. The task mentioned above can be partially accomplished by Instruction-pix2pix (IP2P) \cite{brooks2023instructpix2pix}, which creates a human instruction and corresponding image editing dataset for model training.
However, these techniques cannot simultaneously be highly effective in both global and local editing.

\begin{figure}[tb]
\includegraphics[width=\linewidth]{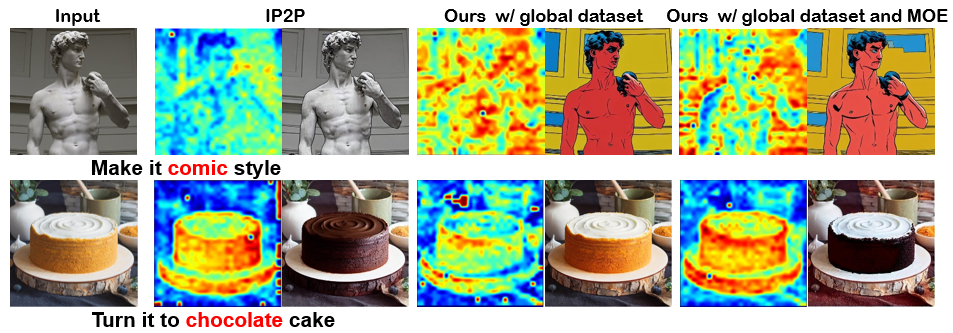}
\caption{\small{Cross-attention heat map of the core words and generated image of different methods.}
}
\label{fig:2}
\end{figure}

We discover through experimental analysis that IP2P performs poorly in some tasks involving the global manipulation of images, as evidenced by the instruction to ``Make it comic style'' in Fig. \ref{fig:2}. The impact of the heatmap of the core words is weak, which is the cause.  This method uses prompt-to-prompt (P2P) \cite{hertz2022prompt} to process feature maps of cross-attention to construct the image editing dataset but is ineffective for transforming global features.
Therefore, we are looking for ways to enhance our capabilities in this area. ControlNet \cite{zhang2023adding} can generate images conditioned on global information extracted from input images, which gives it the ability to perform general image manipulations. Based on this method, we propose to generate a certain amount of global manipulation data.
However, when we test the results of local image editing, we discover that the model's performance is diminished. Through the w/ global dataset method in Fig. \ref{fig:2}, it can be found that the heat map of using the instruction ``Turn it to a chocolate cake'' is lessened, which results in a worse generation result.
In order to solve this problem, considering the instruction semantic distinctions of different image manipulation tasks, we design a Fusion module and MOE model between the text encoder and the diffusion model to discriminate between differences. On the basis of retaining the original ability of the diffusion model to generate text-guided images, it is used to learn the intrinsic mapping of text-guided and image transformation knowledge for different tasks and to achieve multiple image manipulation tasks simultaneously.

To the end, we train a diffusion model equipped with an MOE controller, which can automatically adapt to different expert models under any text instructions and achieve arbitrary manipulation tasks for open-domain images. In Fig. \ref{fig:2}, the w/ global dataset and MOE method produce good results for both global and local manipulation tasks. The examples selected in Fig. \ref{fig:1} demonstrate that our model can perform corresponding transformations for global and local image editing tasks (style transfer, background changing, local editing, entity and attribute swapping) according to user instructions, displaying a strong zero-shot generalization capability. Our contributions are as follows:
\begin{itemize}[leftmargin=*, noitemsep, nolistsep]
\item We generate a large-scale dataset for text-to-image global manipulation, which can guide the model in instruction learning, endowing it with image global manipulation guided by any text instruction.
\item We develop an MOE model that can automatically adapt to different image manipulation tasks when given different text instructions. This model has both global and local image editing capabilities.
\item Numerous experiments show our model's comprehensive superior performance on open-domain image global and local editing tasks as compared with other SOTA methods.
\end{itemize}

\begin{figure*}[ht]
\includegraphics[width=\linewidth, height=0.33\textwidth]{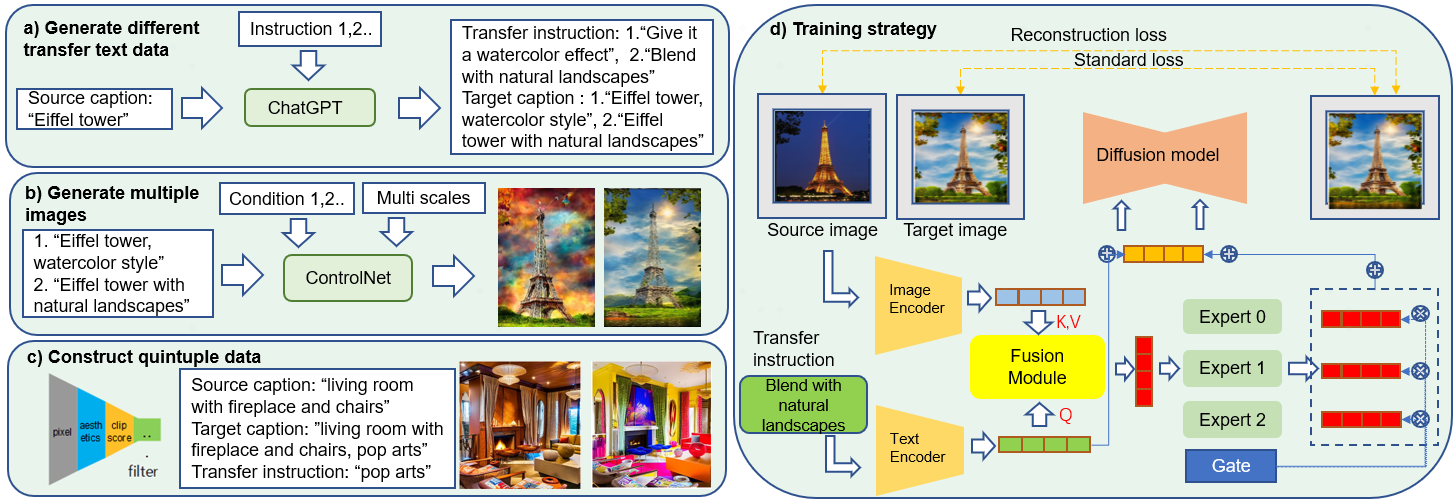}
\caption{\small{An overview of our approach. \textit{Left}: pipeline of dataset construction. \textit{Right}: MOE controller structure.}}
\label{fig:3}
\end{figure*}

\section{Related work}
\label{sec:prior}
{\bf image manipulation.} In the past few years, a large number of works have employed the diffusion model to deal with image manipulation tasks. Besides the text prompt, some approaches require additional inputs, such as user-provided masks to edit local areas of the input images \cite{avrahami2022blended, yang2023paint} and reference images to transfer global styles~\cite{goel2023pair, chen2023artfusion}. Based on the cross-attention control mechanism~\cite{hertz2022prompt}, self-attention control~\cite{wang2023compositional}, segmentation tools~\cite{kirillov2023segment} or DDIM inversion theory, another group of approaches~\cite{tumanyan2023plug, cao2023masactrl, xie2023edit} propose to edit images without user-provided masks. Specifically, \cite{meng2021sdedit} synthesizes realistic images by iteratively denoising through a stochastic
differential equation (SDE) based on a diffusion model generative prior. Recently, personalized image generation methods~\cite{ruiz2023dreambooth, ma2023subject} also synthesize new images based on the reference images to achieve image manipulation with higher degrees of freedom.
However, these image manipulation methods require additional inputs or cannot support arbitrary user instructions, restricting their real-world applications.

{\bf learning instruction.} In terms of learning to follow instructions, 
\cite{ouyang2022training} achieves astonishing results in a wide range of NLP instruction tasks by fine-tuning human instructions. \cite{wang2023context} learns visual language cues by modeling various visual language tasks through in-context learning, achieving text-guided image editing capabilities. InstructEdit \cite{wang2023instructedit} proposes an integration pipeline by employing BLIP2, a language processor, and an image segmentation tool to achieve instruction-guided image editing.
\cite{nguyen2023visual} transforms visual cues into editing instructions, and learning the editing direction of text can enable executing the same editing on new images. The most related work of our method is the IP2P \cite{brooks2023instructpix2pix}, which constructs an instruction-based image editing dataset through P2P \cite{hertz2022prompt} and GPT tools.
However, the aforementioned instruction-based methods do not consider the difference among image manipulation tasks, \textit{e.g.}, global style transfer and local object or attribute swapping. Therefore, we are inspired to design an MOE method to learn human instruction-related global and local image editing tasks, endowing the model with a unified ability.

\section{Methodology}
\subsection{Construct global manipulation dataset}
We create a global transformation dataset using a portion of the high-quality image data from lion-5b \cite{schuhmann2022laion}.
The first step is to input the original image caption and reference examples to ChatGPT and, combined with different instructions and in-context examples, generate corresponding target image captions that conform to the original image caption. This promotes diversity while ensuring cost savings and controllability, as shown in Fig. \ref{fig:3}(a).
We feed the ControlNet model with the aforementioned generated text data in Fig. \ref{fig:3}(b) to produce pairwise global manipulation images.
For batch image generation, we specifically use six image condition extraction methods and various text scale coefficients. Finally, in Fig. \ref{fig:3}(c), we screen the text image quality using a variety of filtering conditions, including image resolutions, aesthetic scores, and CLIP scores, and build a comprehensive global manipulation dataset for model learning.

\subsection{Mixture-of-expert controllers}
Various image manipulation tasks are included in our extensive training dataset; these tasks can essentially be divided into two categories: global and local. Analysis reveals that text instructions and manipulated image can help us understand this difference.
We use a multimodal feature fusion module and a combination of expert controllers to learn the semantics of various text instructions and adapt to various image manipulation capabilities in order to integrate the powerful ability to generate images from text for various image manipulation tasks. Specifically, we add a cross-attention network to fuse text and image features, expert models with different abilities, and a gated system to automatically adapt to different image manipulation tasks, as shown in Fig.~\ref{fig:3}(d).
We conducted experiments with different numbers of experts and found that more expert systems have limited improvement in image transformation effects. So we intuitively divide the image transformation tasks into three main categories: global editing, local editing, and fine-grained editing, and use three expert models for design. 

Finally, a fusion module, three expert models, and a gate system are created. In the fusion module, input $Q$ is text features, and input $K$ and $V$ are image features generated by CLIP Image Encoder, which can fully integrate text instructions and image transformation. The multi-expert model can be regarded as the multi-head mechanism of the transformer. Each expert model is a two-layer feed-forward network, and the gating model, which is made up of a linear layer and a softmax layer, has an output dimension equal to the number of expert models. Additionally, we examined the text instructions that the three expert models had adapted and discovered that Expert 1 is appropriate for fine-grained local image translation tasks, such as ``change the color to yellow'', Expert 2 is appropriate for global image style transfer tasks, such as ``turn it into cartoon style'', and Expert 3 is appropriate for local image editing tasks, such as ``add fireworks into the sky''. We present the final loss function as follows:

\begin{equation}
\mathcal{L}\!=\!\mathbb{E}_{z,\varepsilon\sim\mathcal{N}(0,I),t} \left\|\varepsilon\!-\!f_\theta(z_{t,tgt},t,c) \right\|_2^2 \!+\! w\!*\!\left\|\varepsilon\!-\!f_\theta(z_{t,src},t,c) \right\|_2^2
\label{eq:1}
\end{equation}
\vspace{-1em}
\begin{equation}
where \quad c=\sum_{i=1}^{n}g_i(x)f_i(x)+E_t(y_t), 
g_i(x) = softmax(W_{i}x),
\label{eq:2}
\end{equation}
\vspace{-1em}
\begin{equation}
\label{eq:3}
\begin{aligned}
    &{CA}(Q, K, V) = {softmax}\left(\frac{QK^T}{\sqrt{d}}\right)\cdot V \\
    \textit{where} \quad &Q = W_Q \cdot E_t(y_t), K = W_K \cdot E_i(y_i), V = W_V \cdot E_i(y_i).
\end{aligned}
\end{equation}

Eq. (\ref{eq:2}) is a simplified expression of the mixture-of-expert model, where $c$ is the text condition for the input diffusion model. By taking the fusion feature $x$ as input, Eq. (\ref{eq:2}) indicates that the modified condition $c$ is a linear combination of the output of all $n$ experts: $f_i (i=\{1,\cdots,n\})$ added by $E_t(y_t)$. As shown in Eq.~(\ref{eq:2}), $g_i(x)$ is the softmax output of the gate, $W_i$ is the weights of the linear layer.
Eq. (\ref{eq:3}), The $CA(Q,K,V)$ represents the cross-attention network within the fusion module. $E_i$ and $E_t$ correspond to the CLIP image encoder and text encoder, respectively. The term $y_i$ denotes the source image, while $y_t$ signifies the input text instruction. 

\subsection{Reconstruction loss}
Our global transformation dataset is constructed through ControlNet. Since ControlNet generates images from the extracted conditions from the original images, it may cause changes in the generated image and the original image entities. To mitigate this issue, as shown in Fig. \ref{fig:3}(d), we follow the idea of Dreambooth \cite{ruiz2023dreambooth} by adding a reconstruction loss related to the original image during the training phase to constrain and ensure the consistency of the image entities. The complete loss function is shown in Eq. (\ref{eq:1}), where the subscript $src$ refers to the original image, $tgt$ refers to the target image, $\theta$ represents the model parameters to be learned, $t$ is the diffusion time step, and $w$ is the weight coefficient of the image reconstruction loss.

\section{Experiments}
\subsection{Qualitative Comparison}
\label{qualitative}
\begin{figure}[ht]
\includegraphics[width=\linewidth]{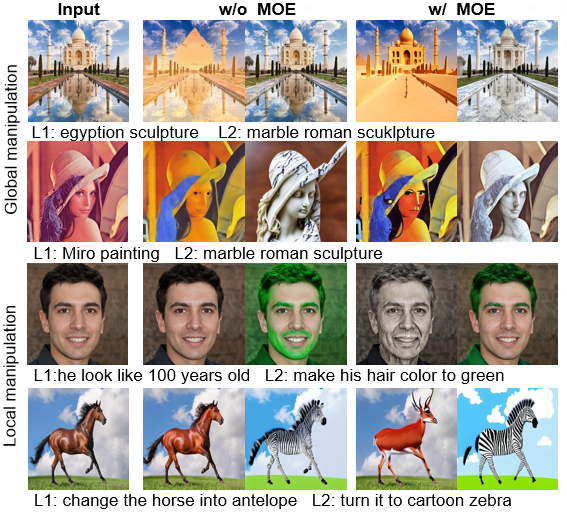}
\caption{\small{Comparison of image global and local manipulation tasks with and without mixture-of-experts.}
}
\label{fig:4}
\end{figure}

\begin{figure*}[h]
\includegraphics[width=\linewidth]{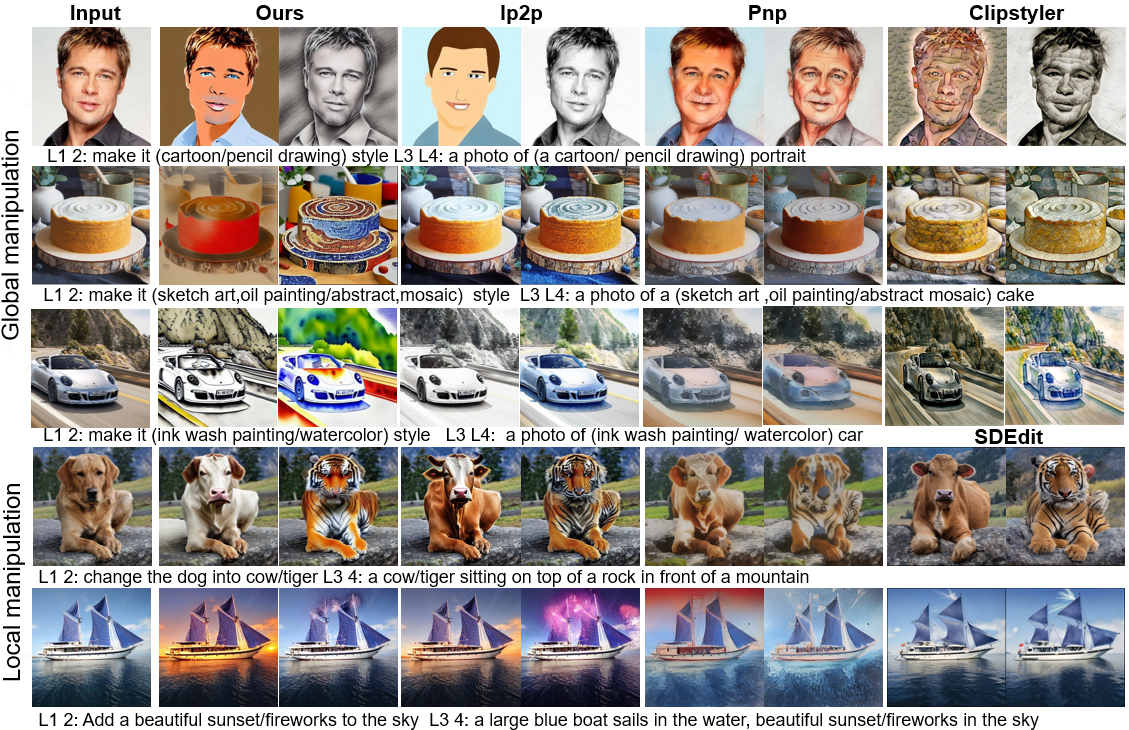}
\caption{\small{Global and local image manipulation qualitative comparison experimental results.}}
\label{fig:6}
\end{figure*}

In the tasks of global and local editing on images, our method is qualitatively compared with PNP \cite{tumanyan2023plug}, IP2P \cite{brooks2023instructpix2pix}, Clipstyler \cite{kwon2022clipstyler} and SDEdit \cite{meng2021sdedit}.
In the global manipulation of Fig. \ref{fig:6}, it can be found that Clipstyler is somewhat lacking in the ability to transfer styles in the open domain. PNP slightly changes the object details in the target images, \textit{e.g.}, the face and cake details, and IP2P's performance is poor for complex style transfer tasks. Our method performs the best in these complex style transfer tasks.
The local manipulation in Fig. \ref{fig:6} demonstrates that our method outperforms SDEdit and PNP while coming close to matching IP2P in terms of performance. Images generated by SDEdit and PNP have unsuccessful editing or minor blurriness issues. Furthermore, when IP2P is jointly trained with the global manipulation dataset we constructed and the original dataset in \cite{brooks2023instructpix2pix}, it exactly corresponds to the w/o MOE method on the left side of Fig. \ref{fig:4}. However, the performance of local editing is not as good as the w/ MOE method on the right side of Fig. \ref{fig:4}. 

\begin{figure}[ht]
\includegraphics[width=\linewidth]{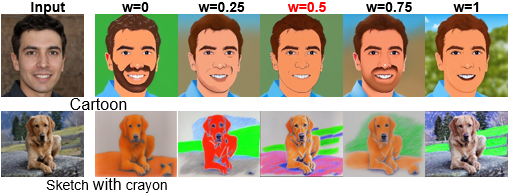}
\caption{\small{The effect of image style transfer with different w.}}
\label{fig:5}
\end{figure}

\begin{table}[tb]
\centering
\resizebox{\linewidth}{!}{
\begin{tabular}{c|ccccc}
\hline
   &  \multicolumn{2}{c}{Global manipulation}  & 
 \multicolumn{2}{c}{Local manipulation}  &
 User study\\ 
Methods & CLIP-T $\uparrow$ & CLIP-D $\uparrow$ & CLIP-T $\uparrow$ & CLIP-D $\uparrow$  &
PREFER SCORE $\uparrow$ \\ 
\hline
Ours w/ MOE  & {\bf 0.1548}  &  {\bf 0.2817}  & 0.2953 & {\bf 0.2687} & {\bf 0.3346} \\
Ours w/o MOE & 0.1489  &  0.2774  & 0.2614 & 0.2535 & - \\ 
\hdashline
IP2P         & 0.124   &  0.2715  & {\bf 0.3007} & 0.2657 & 0.2572 \\
PNP          & 0.1015  &  0.2677  & 0.1883 & 0.2541 & 0.142 \\
Clipstyler   & 0.1531  &  0.2246  & -      & -      & 0.1833 \\
SDEdit       & 0.0833  &  0.2678  & 0.2037 & 0.2499 & 0.0829 \\
\hline
\end{tabular}%
}
\caption{\small{Quantitative comparison results.}}
\label{tab:1}
\end{table}

\subsection{Quantitative Comparison}
\label{quantitative}
We select 100 text instructions and 50 images to conduct comparative experiments with PNP \cite{tumanyan2023plug}, IP2P \cite{brooks2023instructpix2pix} Clipstyler \cite{kwon2022clipstyler} and SDEdit \cite{meng2021sdedit} on global and local image manipulation tasks. The employed evaluation metrics are the similarity between target and generated images, \textit{i.e.}, CLIP-T and the directional consistency of text-image pairs, \textit{i.e.}, CLIP-D \cite{brooks2023instructpix2pix} calculated by the CLIP ViT-L/14 model.
When compared to other methods, our results achieve state-of-the-art results in terms of global manipulation, as shown in Table \ref{tab:1}.
In terms of local editing, our results achieve SOTA results on the CLIP-D metric. By comprehensively considering the two indicators, we can find that our method and IP2P have comparable performance on local editing.
We conducted a user study to further test the effectiveness of our experiments, with a total of 10 participants. Based on the dataset for quantitative comparisons, we randomly shuffled the images and asked the participants to rank the generated results according to image quality and semantic matching degree. The average proportion of participants who rank the corresponding generated images as the top 1 is used to calculate each method's final score.
Statistics in Table \ref{tab:1} reveal that our method performs best in a variety of image manipulation tasks and obtains a score of 0.3346, which is higher than that of other approaches.

\subsection{Ablation Study}
{\bf Mixture-of-expert model.} The comparison of the effects of using the MOE method or not on global and local image manipulation tasks is shown in Fig. \ref{fig:4}. The model's ability to perform local editing is found to be significantly improved when MOE is used for training, and the generated effect on global image manipulation is found to be more consistent with the input image content and text instructions, proving that the addition of the MOE method improves a variety of image manipulation tasks. The CLIP scores in Table \ref{tab:1} also confirm that the w/ MOE method performs both global and local image manipulations more effectively than the w/o MOE method.

{\bf Reconstructing loss.} Fig. \ref{fig:5} depicts the subtle impact of adding different reconstruction loss coefficients on the generated results. When $w=0$, the reconstruction loss is not used for training constraints, and the generated images will show a certain degree of distortion. The ``dog's crayon sketch transfer'' with $w=1.0$ retains too much content from the original image and loses some translation ability. When $w$ is between 0.25 and 0.75, the fidelity of the image and the transformation effect achieve a relatively good balance, and finally, $w=0.5$ is used in our model training.

\section{Conclusion}
To address the issue of text-conditioned controlled image generation, we propose MoEController, which primarily deals with global and local manipulation tasks of images based on human instructions. In constructing the training data, we integrate the capabilities of the large language model ChatGPT and the conditional image generation model ControlNet with Stable diffusion. In model training, we use multi-expert models to adapt the capabilities of various semantic instructions corresponding to different image manipulations. Extensive experiments show that our model achieves state-of-the-art results in global and local image manipulation tasks. Moreover, our MoEController can be easily extended to more general human instruction input and a wider variety of image manipulation tasks.

\bibliographystyle{IEEEbib}
\bibliography{main}

\end{document}